\documentclass[a4paper, 12pt, twoside]{article}
\input{JADTstyle.cls}
\usepackage{array}
\usepackage{booktabs}
\usepackage{caption}
\usepackage{graphicx}
\usepackage{tabularx}  
\usepackage{ragged2e}  
\usepackage[none]{hyphenat}
\usepackage{dirtytalk}
\usepackage{csquotes}

\title{\fontsize{18}{21.6}\textbf{The French Drama Revolution: Political Economy and Literary Production, 1700-1900}}

\author[*]{Thiago Dumont Oliveira}

\affil[*]{Centre for Digital Humanities, University of Tartu --- thiago@ut.ee}
\date{\vspace*{-3em}}

\begin{document}

\maketitle

\fancyhead[CE]{\textsc{\footnotesize{Thiago Dumont Oliveira}}}
\fancyhead[CO]{\textsc{\footnotesize{The French Drama Revolution}}}

\selectlanguage{english} 

\setlength{\parindent}{\z@}
\setlength{\baselineskip}{11pt}


\vspace*{2pt}
\abstract*{Abstract}

\begin{footnotesize}{This paper investigates the changing nature of French drama between 1700-1900 using Latent Dirichlet Allocation and Jensen-Shannon Divergence. Results indicate that the topical distribution of French drama changed profoundly after the French Revolution, particularly between 1789 and 1850. Bourgeois themes emerged among the most prevalent topics since the late 18th century. To assess the coevolution of drama and economic growth, I plot the yearly prevalence of topics alongside French GDP between 1700-1900, and discuss these changes in light of the political and economic changes prompted by the French Revolution and the industrialization of the country.  
}\end{footnotesize}

\vspace*{14pt}{\textbf{Keywords:}} \begin{footnotesize}{Computational Literary Studies, Political Economy, Cultural History, Latent Dirichlet Allocation, Jensen-Shannon Divergence.}\end{footnotesize}


\setlength{\parskip}{6pt}
\setlength{\baselineskip}{14pt}

\section{Introduction}

This paper approaches the evolution of French drama from a \textit{longue durée} perspective \citep{braudel1958histoire}, examining literary production through the lens of political economy. I assess which topics grew in importance within French drama between 1700-1900, and discuss these changes in light of the political and economic changes prompted by the French Revolution and the industrialization of the country, which drastically altered French society \citep{hobsbawmrevolution,hobsbawmcapital}. Using Latent Dirichlet Allocation and Jensen-Shannon Divergence, I show that the topical distribution of French drama changed profoundly during the ‘Age of Revolution’, as Hobsbawm dubbed the period 1789-1848. Topics containing bourgeouis themes emerge among the most prevalent topics since the late 18th century and especially after 1820, with the acceleration of French industrialization.  

I use French drama as a window of investigation into the relationship between literature and the social structure. Epistemologically, I take the position that literary production is embedded in political-economic relations \citep{granovetter1985economic}, yet not fully determined by material conditions, since individuals simultaneously shape and are shaped by institutions \citep{agassi1975institutional}. The interplay between drama and the political-economic world can thus be approached as an instance of a question long debated among philosophers of science regarding the relationship between agent and structure \citep{oliveira2018nature}. I concur with \cite{mandelbaum1955societal} that individual action cannot be understood without reference to \say{societal facts}: \say{the actual behaviour of specific individuals
towards one another is unintelligible unless one views their behaviour in terms
of their status and roles, and the concepts of status and role are devoid of
meaning unless one interprets them in terms of the organization of the society
to which the individuals belong} (307-308). My position is that literature is deeply rooted in political-economic relations (or \say{societal facts}, to use Mandelbaum's expression), yet literary production is best approached as \say{total social facts} \citep{mauss1925gift}, and hence not ultimately reducible to political and economic material realities.   
 
Topic modeling has been increasingly used within computational literary studies. \cite{schoch2021topic} uses Latent Dirichlet Allocation (LDA) to analyze French Drama of the
Classical Age and the Enlightenment using plays from 1630 to 1789. \cite{min2019modeling} use Non-Negative Matrix Factorization (NMF), sentiment analysis, and network analysis to study Victor Hugo’s \textit{Les Misérables}. \cite{dahllof2019faces} combine gender studies and Swedish
literary history, using topic modeling to perform a gendered thematic analysis of Swedish classics (1821–1941) and recent bestsellers
(2004–2017). \cite{ginn2024historia} compare LDA and NMF and BERT-type models to examine Roman literature from the founding of Rome to its
fall. \cite{martinelli2024exploring} use neural topic modeling to examine Classical Latin literature. My paper contributes to this literature by drawing on economic history, GDP data, and political economy to discuss the main shifts observable in French drama throughout the 18th and 19th centuries in light of the broader political-economic changes that transformed the country.


\section{Data and Methods}

My data are 1215 French plays published between 1700 and 1900. The corpus was built using the French-language corpora made available by the DraCor project \citep{fischer2019programmable} and it was retrieved via their API (\url{https://dracor.org/doc/api}). Documents were processed using spaCy's French language model  and normalized through lemmatization because French is a highly inflected language. Preprocessing used a part-of-speech filter to retain only nouns, proper nouns, verbs, and adjectives. French stopwords were removed, as well as highly frequent light verbs (e.g. être, avoir, faire, dire) and address terms (e.g. monsieur, madame). In order to assess the changing nature of French Drama in the 18th and 19th century I use Latent Dirichlet Allocation (LDA) and Jensen-Shannon Divergence (JSD). 

LDA is a machine learning method widely used in topic modeling to identify the latent semantic structure of texts. LDA is a generative probabilistic model that models documents as probability distributions over latent topics, while topics are modeled as probability distributions over the words of the vocabulary (\citet{blei2003latent}, \citet{steyvers2007probabilistic}, \citet{blei2012probabilistic}). Instead of reading \textit{between} the lines to uncover authors’ purposes, LDA shifts the analysis to reading \textit{above} the lines, such that the topical structure of the texts is derived from the entire constellation of documents to which they belong. It is thus a powerful instrument in approaching the coevolution of literature and economics from a \say{distant reading} perspective, rather than focusing on the exegesis of a few canonical plays \citep{moretti2013distant}. Moreover, LDA is particularly well suited for my purposes, as it is an effective tool for exploring questions that bridge literary studies and socioeconomic perspectives on culture \citep{dimaggio2013exploiting}.

One of the crucial issues with LDA is the selection of the number of topics, a hyperparameter of the model which must be decided apriori. The optimal number of topics depends on the characteristics of the corpus, on the number of tokens per document, and on the level of granularity adequate to the research question (\citet{rhody2012topic}, \citet{jockers2013macroanalysis}, \citet{sbalchiero2020topic}). Although there are numerous studies on the detection and selection of the number of topics (\cite{griffiths2004finding}, \cite{newman2010automatic}, \cite{roder2015exploring}, \cite{gan2021selection}), the optimal number of topics obtained via metrics such as perplexity and topic coherence is hardly a good benchmark,  since the appropriate number of topics ultimately depends on topic interpretability and on the level of granularity required by the problem at hand \citep{wehrheim2019economic}. Thus, while LDA is an unsupervised machine learning technique, historical analysis and a theoretically informed understanding of the object of study remain essential in evaluating the coherence of topics and in selecting the number of topics \citep{mohr2013introduction}.

   My choices regarding preprocessing and the number of topics were based on topic interpretability and coherence after experimenting with different POS strategies and number of topics. Since my main goal is to trace the rise of economic discourse in French Drama, a 10-topic model proved sufficient in yielding topics with a strong presence of economic discourse and bourgeois everyday life themes, as well as topics that are reminiscent of Greek tragedy and the aristocratic zeitgeist. As a robustness check I also used NMF for the topic modeling analysis, which yielded similar results (see the Appendix). 

   In order to identify periods of rapid topical change in French Drama I use Jensen-Shannon divergence (\cite{lin2002divergence}, \cite{nielbo2023pandemic}). JSD measures how much the topic distribution of documents changes between 2 consecutive years. It is defined by

\[\mathrm{JSD}(P \mid Q)=\frac{1}{2} \mathrm{D}(P \mid M)+\frac{1}{2} \mathrm{D}(Q\mid M)\]

where \( M = \frac{1}{2}(P + Q) \) and \( D \) is the Kullback--Leibler divergence: 

\[D(P \mid Q) = \sum_{i : P_i > 0} P_i \log_2 \frac{P_i}{Q_i}\]

Lastly, I use multidimensional scaling to build a semantic map from the cosine distance between documents’ topic vectors. Three labels were added on the map, corresponding to three topics of particular interest to my argument, which I further explore throughout the paper. The labels were added at the centroid of the five documents with the highest topic proportions for each topic of interest. These five representative works per topic are listed in Table 1.

\section{Results}

Table 1 shows the average prevalence of topics between 1700 and 1900, the variation in prevalence over the period ($\Delta_{1700\text{--}1900}$), the 15 most important words, and 5 representative works per topic. $\Delta_{1700\text{--}1900}$ measures the change in topic prevalence over the period. It was obtained by calculating the slope of an OLS regression for each topic and multiplying it by 200, yielding the overall change of the fitted line over the period. 

Topics 0 and 1 barely change over time and their slope is not significantly different from 0. Topics 2, 3 and 7 are statistically different from 0, but their coefficients are very small. Moreover, these topics are among the least prevalent topics in my table. Topic 6 is an intermediate case ($\Delta_{1700\text{--}1900}$ = -- 0.13). While its average prevalence is low (0.07),  it was one of the most important topics in the early 18th century, although not as important as topics 5 and 8 (see Figure 1, which plots the evolution of the 10 topics over the period). 

Topics 4 and 9 stand out as \say{hot topics} ($\Delta_{1700\text{--}1900}$ = + 0.37 and + 0.23)., whereas 5 and 8 are the \say{cold topics} ($\Delta_{1700\text{--}1900}$ = -- 0.24 and -- 0.27). The aggregate prevalence of topics 5 and 8 falls from 0.5 in the first decade of the 18th century to 0.07 in the last decade of the 19th century. On the other hand, the aggregate weight of topics 4 and 9 increases from 0.11 to 0.51, making them the most prevalent topics in the late 19th century. 

\begin{figure}[!htbp]
\centering
\includegraphics[scale=0.6]{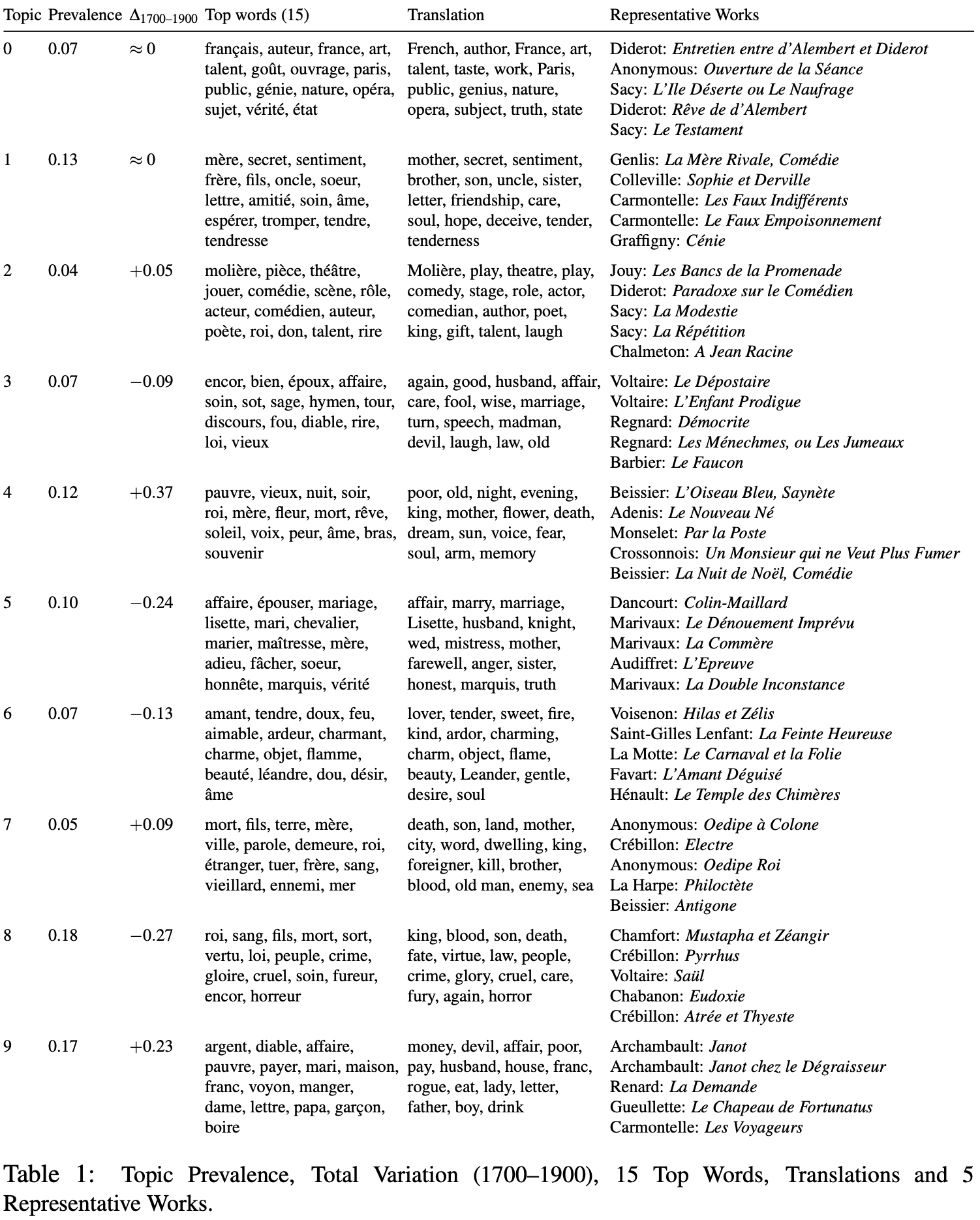}
\end{figure}

\newpage
Three topics are of particular interest to my research question, namely topics 4, 8 and 9. Figure 1 shows that topics 4 and 9 grew significantly in importance after 1800, becoming the two most important topics by the end of the century, whereas topic 8 lost considerable weight after 1800. For the ease of interpretation, I label these topics as \textit{Bourgeois Life} (topic 4), \textit{Aristocratic Life} (topic 8) and \textit{Household Economics} (topic 9). Topic 4, and especially topic 9, have a clear material dimension and fall under the umbrella of bourgeois comedies. I chose \textit{Household Economics} as a label for topic 9 because of the presence of words such as \textit{money}, \textit{poor}, \textit{pay}, \textit{husband}, \textit{house} and \textit{franc} among its top 15 words. While the economic dimension is less explicit looking at the top 15 words of topic 4 (except for the word \textit{poor}), an analysis of its representative works clearly indicates that this topic is mostly about the institutions and the social reproduction of the bourgeois family. Topic 8, on the other hand, focuses on questions of virtue, justice, and political authority which are reminiscent of the absolutist state. The top words of this topic include \textit{king}, \textit{virtue}, \textit{law}, \textit{people}, \textit{crime}. This topic thus touches on issues regarding the political organization of the state, in modern parlance, that were central to political philosophers from Plato's \textit{Republic} and Aristotle's \textit{Politics} to Machiavelli's \textit{Prince} and Hobbes's \textit{Leviathan}. 

\begin{figure}[!htbp]
\centering
\includegraphics[scale=0.5]{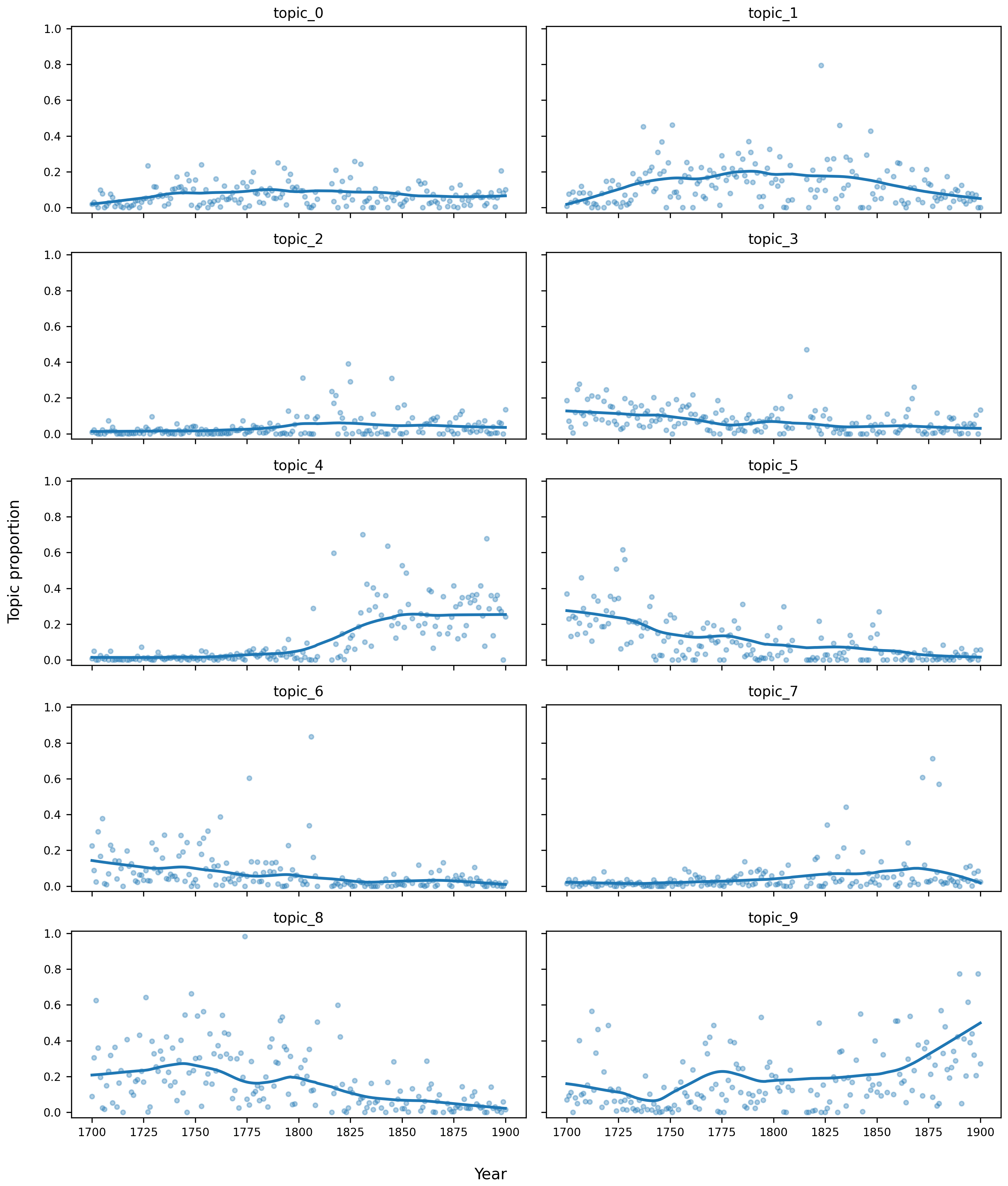}
	\caption{Temporal evolution of the topics}
\end{figure}

Figure 2 shows the evolution of these three topics vis-à-vis French GDP.\footnote{I use real GDP per capita in 2011 US dollars. The data were retrieved from the Maddison Project and are available from \url{https://www.rug.nl/ggdc/historicaldevelopment/maddison/}} The decreasing importance of \textit{Aristocratic Life} since the 1750s seems to reflect the rise of anti-monarchic sentiment during this period, inspired by Montesquieu's \textit{The Spirit of Laws} (1748), Rousseau's \textit{The Social Contract} (1762), and Diderot \& D'Alembert's \textit{Encyclopédie} (1751-1772). 

There is a striking similarity between the evolution of GDP and \textit{Household Economics}. Both series grew very slowly during the 18th century (although \textit{Household Economics} oscillated more than GDP), and both series took off between 1825-1850. GDP growth rates increased considerably after 1825 as France initiated its industrialization, with the first railway being completed in 1827 (Saint-Étienne to Andrézieux). 

The development of French capitalism thus marks a double shift in literary production: from tragedies to comedies, and from topics that reflect on the political authority of rulers to more mundane themes, reflecting the material reality of the nascent bourgeoisie. Continued growth and prosperity throughout the rest of the century, coupled with the proclamation of the Third Republic in 1870, completely shifted the politico-economic milieu in which authors and readers are socially embedded. The series \textit{Bourgeois Life} is also essentially flat for most of the 18th century, but it considerably grows in importance between the French Revolution and 1850, remaining constant thereafter. Hence, this topic seems less influenced by the economic circumstances than \textit{Household Economics}, and its evolution mirrors instead the changing political context prompted by the French Revolution and the novel political ideas originating from the French Enlightenment.

\begin{figure}[!htbp]
\centering
\includegraphics[scale=0.3]{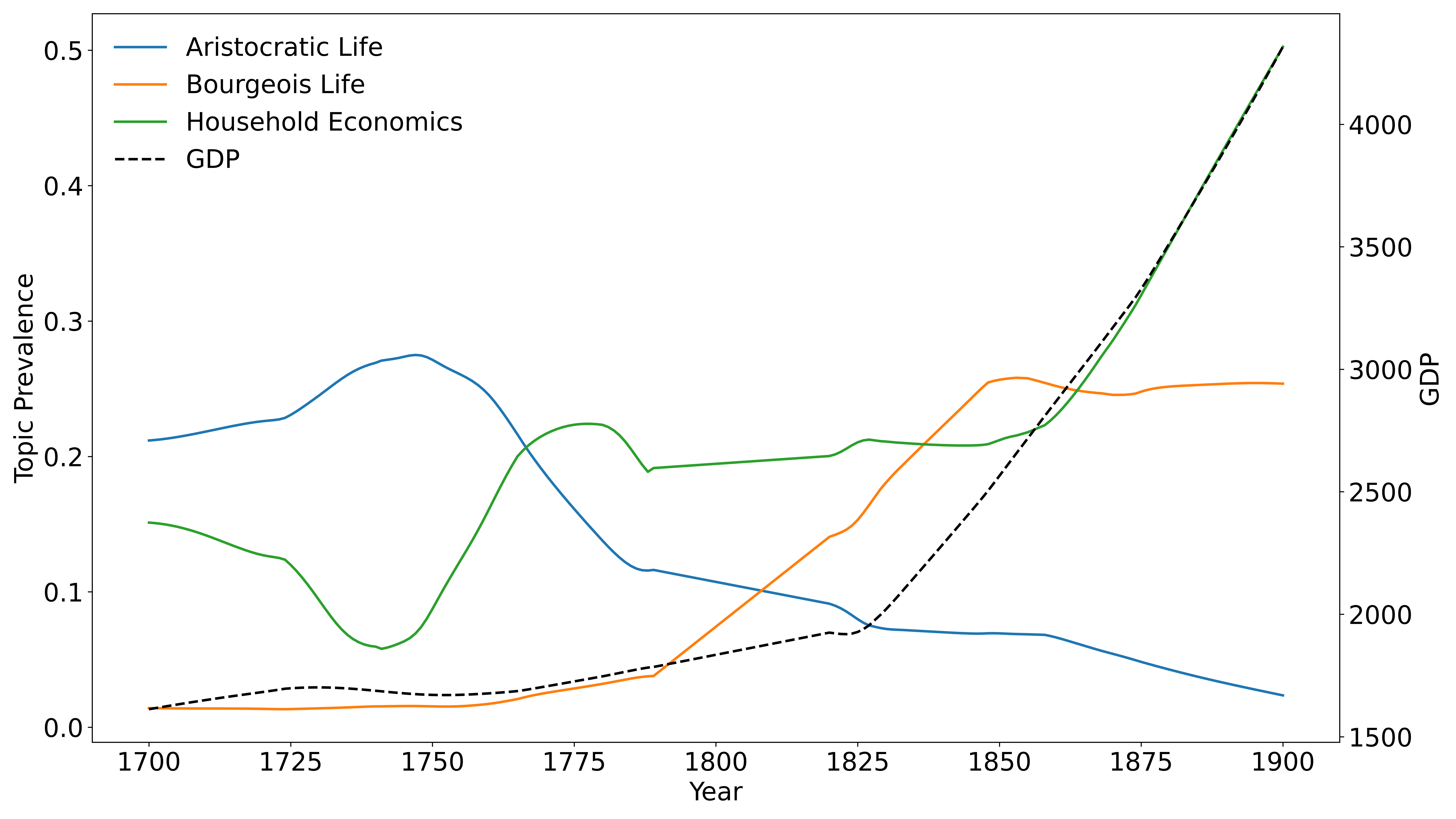}
	\caption{Topic Prevalence vs Real GDP per capita in 2011 US\$}
\end{figure}

The growing importance of these two series map rather neatly into Hobsbawm's periodization in \textit{The Age of Revolution: 1789–1848} and \textit{The Age of Capital: 1848–1875} \citep{hobsbawmrevolution,hobsbawmcapital}. While \textit{Bourgeois Life} grows in importance during the Age of Revolution, but flattens during the Age of Capital, \textit{Household Economics} would only increase in importance in the 1850s, precisely when Hobsbawm identifies the consolidation of capitalism. As French capitalism developed and a bourgeois class emerged in the country, the nature of French drama changed drastically. Jensen-Shannon Divergence shows that the textual profile of French drama quickly changed in the first half of the 19th century (Figure 3). The peak in the JSD series reflects the concomitant decrease in importance of \textit{Aristocratic Life} and increase in \textit{Bourgeois Life} between 1800-1850. 

\begin{figure}[!htbp]
\centering
\includegraphics[scale=0.3]{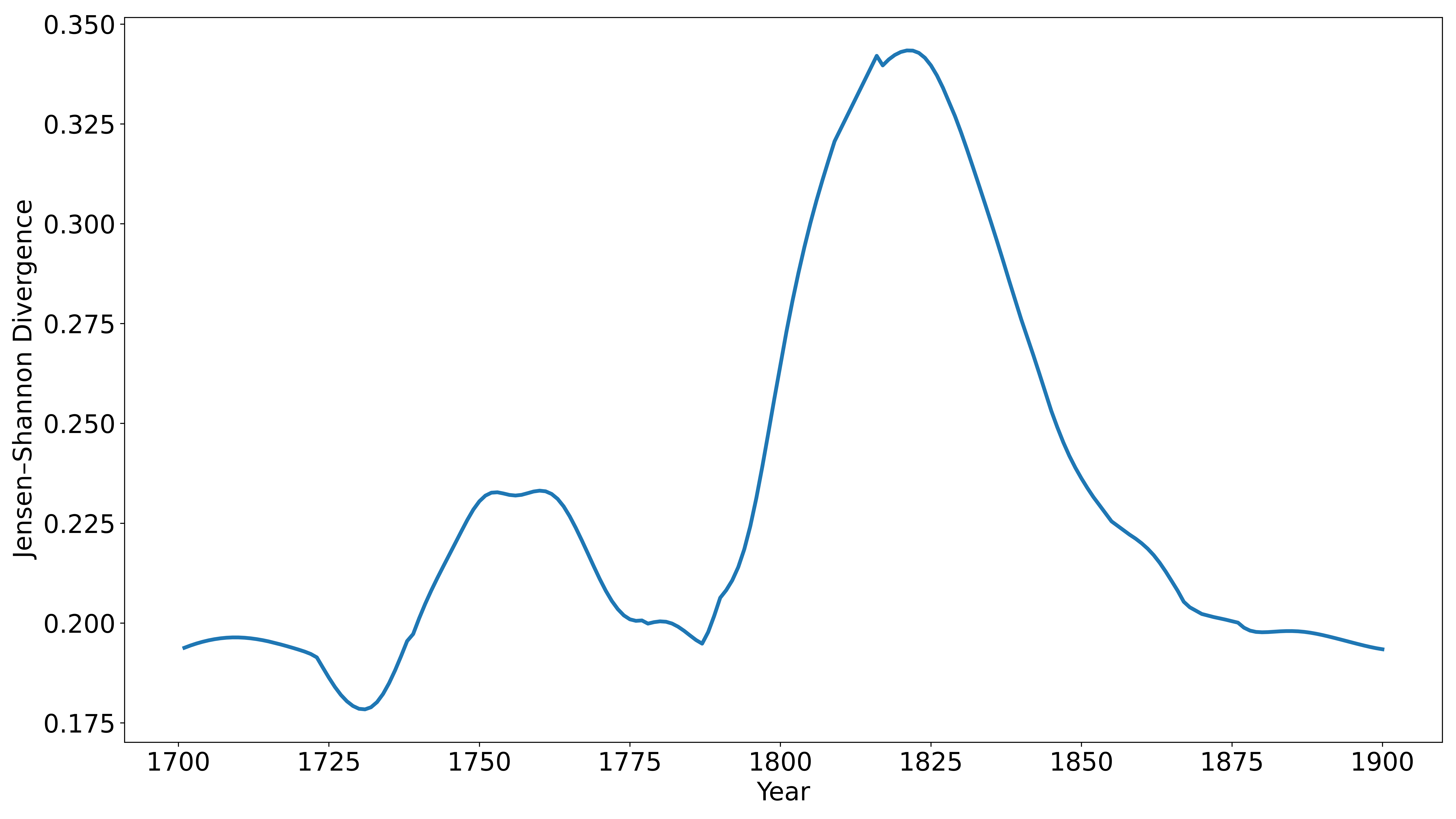}
	\caption{Jensen-Shannon Divergence}
\end{figure}

French plays shifted toward economic discourse and away from discussions of sovereignty, virtue and violence. This can also be seen by the fact that plays from the second half of the 19th century cluster in the lower left region of the semantic map, indicating that they share a textual profile that distinguishes them from plays from the 18th century (Figure 4). 

\begin{figure}[!htbp]
\includegraphics[scale=0.3]{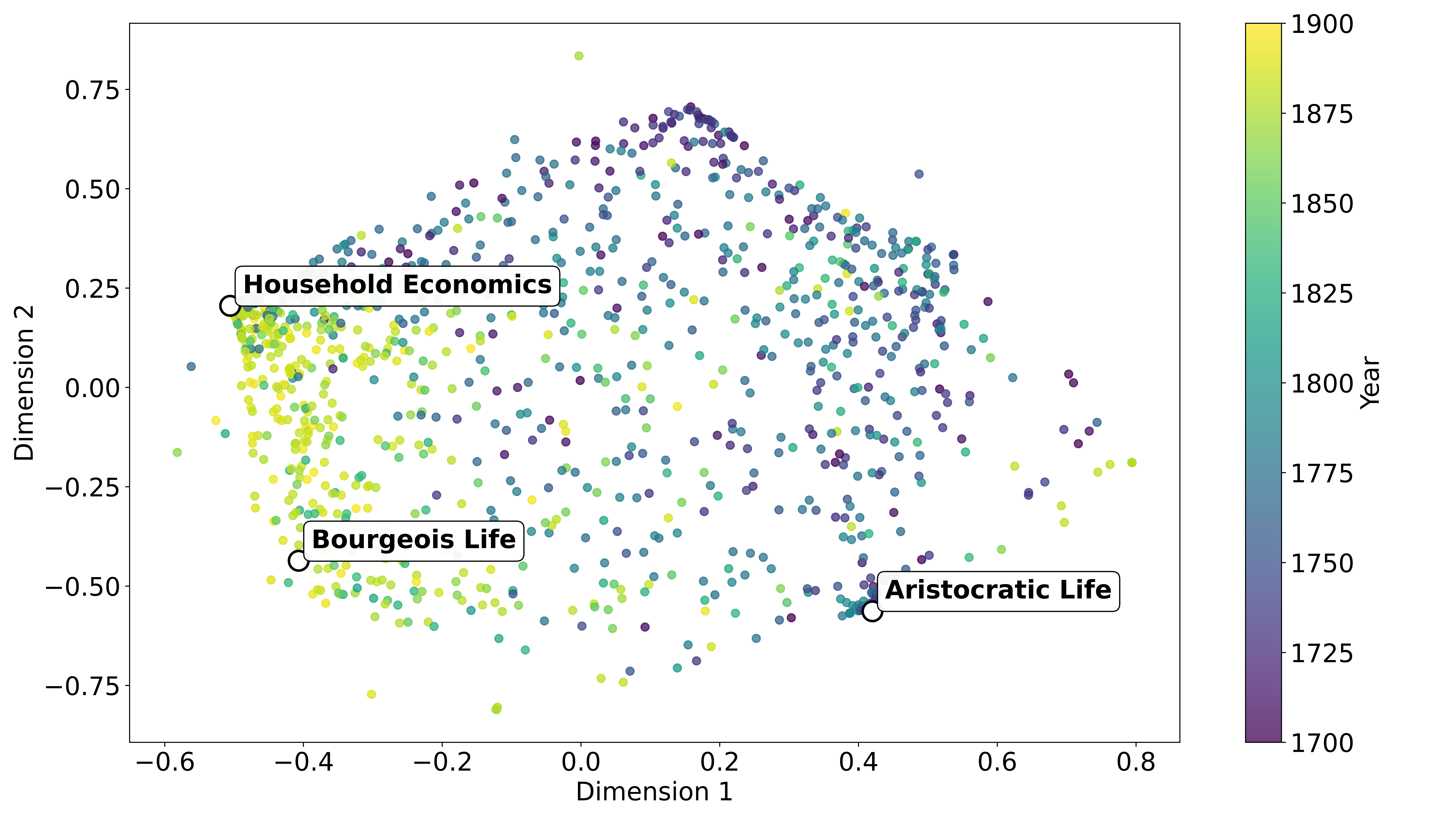}
\centering
	\caption{Semantic Map, MDS of cosine distance between LDA topic distributions. The labels are at the centroid of the 5 representative documents.}
\end{figure}

\section{Conclusion}

This paper has investigated the changing nature of French drama between 1700 and 1900, situating literary production within the broader political-economic
transformations which utterly reconfigured French society – and drama, as my results
indicate. Applying LDA and JSD to a corpus of 1215 French plays, I have shown that the topical distribution of French drama changed profoundly between the late eighteenth and the mid-nineteenth centuries. In particular, the topic \textit{Aristocratic Life}, which emphasizes sovereignty, virtue, law, and political authority, declined steadily in prevalence over time, while topics \textit{Bourgeois Life} and \textit{Household Economics} gained prominence after the French Revolution. 

The thematic profile of French drama radically changed as the absolutist state weakened and capitalist social relations developed. Themes centered on political authority and aristocratic values gave way to more mundane, bourgeois themes, centered on everyday life, domestic relations, and economic concerns. The striking similarity between the series \textit{Household Economics} and \textit{GDP}, and more broadly the coevolution of literary production and economic development, deserve further investigation. More generally, this study illustrates how topic modeling can be used to trace large-scale changes in literary discourse over time, providing a quantitative complement to historical and theoretical approaches to the relationship between literature and political economy.

\section*{Acknowledgments} 

Funded by European Union project "The Center for Digital Text Scholarship" under 566 grant agreement ID 101186601.

\bibliographystyle{plainnat}
\bibliography{references}         

\section*{Appendix}

As s a robustness check to the LDA analysis, I use Non-Negative Matrix Factorization (NMF), a deterministic linear-algebraic method widely used in topic modeling (\cite{lee1999learning}, \cite{lee2000algorithms}). Table 2 shows the 10 topics obtained using NMF. While there are some interesting differences, the main results remain the same. 

Notice that topic 3 is very similar to the topic \textit{Household Economics} previously obtained. Not only do their the top words largely overlap (\textit{money}, \textit{franc}, etc), but also its average prevalence (0.12) and its prevalence variation (+0.37) resemble that of \textit{Household Economics}. The main difference is that this topic also incorporates words such as \textit{poor}, which were part of topic \textit{Bourgeouis Life}. Hence my two previous topics that reflect the rise of the bourgeois class were combined into a single topic when using NMF instead of LDA. 

My other topic of interest, \textit{Aristocratic Life}, also finds a counterpart in topic 1 on Table 2. Notice the presence of words such as  \textit{crime}, \textit{people}, \textit{virtue}, \textit{law}, and \textit{tyrant}. Its average prevalence (0.11) and its prevalence variation (-0.17) indicate that this topic was relatively important in the 18th century, but lost importance in the 19th century, similarly to our previous results regarding the topic \textit{Aristocratic Life}. 

\begin{figure}[!htbp]
\centering
\includegraphics[scale=0.6]{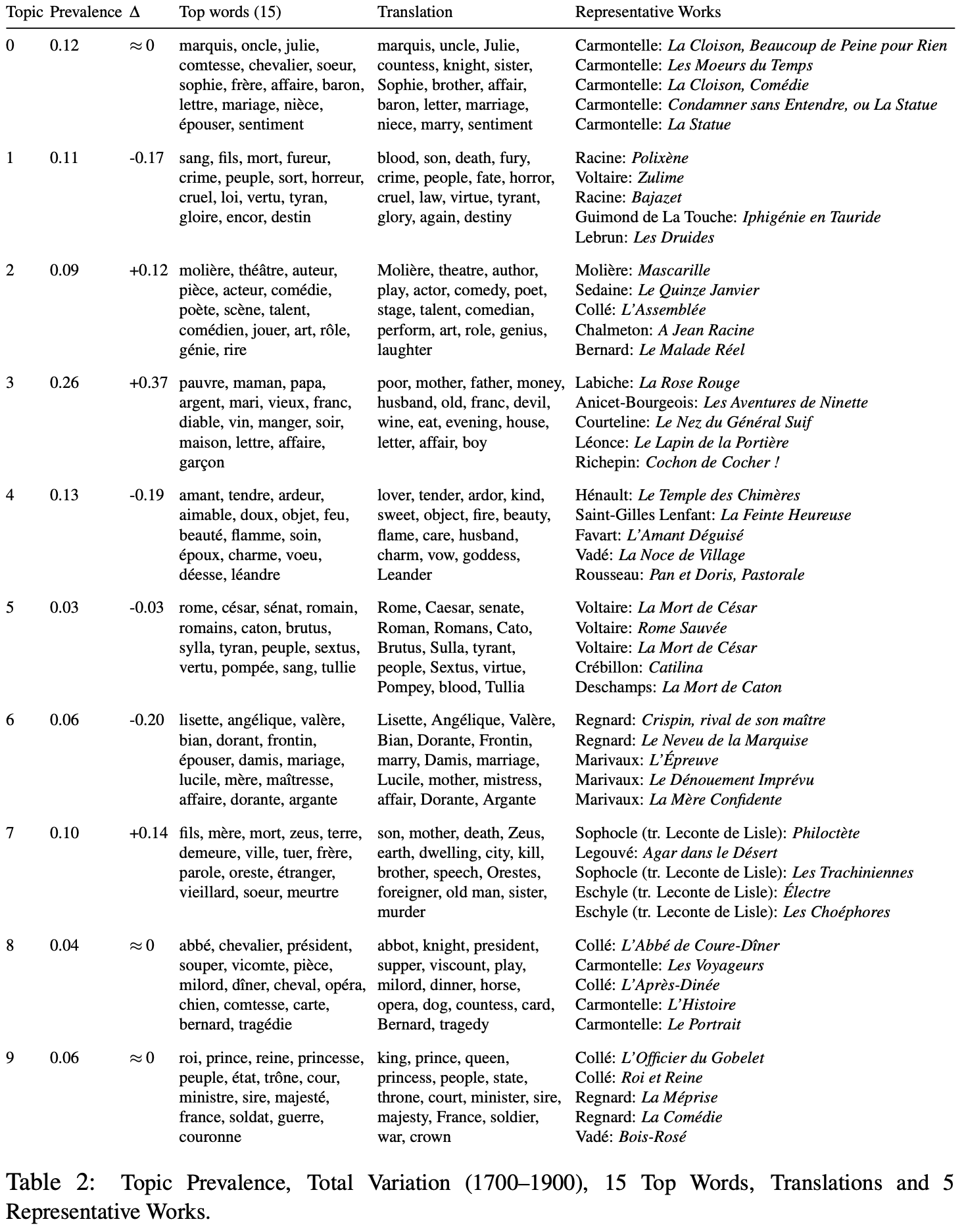}
\end{figure}

\end{document}